\documentclass[11pt,letterpaper]{article}
\usepackage{emnlp2017}
\usepackage{times}
\usepackage{latexsym}
\usepackage{amsmath}
\usepackage{tabularx}
\usepackage{graphicx}
\usepackage{linguex}
\usepackage{multirow}
\usepackage{array}
\usepackage{booktabs}

\emnlpfinalcopy

\def\emnlppaperid{***}

\newcolumntype{x}[1]{>{\centering\arraybackslash\hspace{0pt}}p{#1}}

\title{Towards Linguistically Generalizable NLP Systems:\\ A Workshop and Shared Task}

\newcommand{\affilumdling}{${}^{\triangle}$}
\newcommand{\affilumdcs}{${}^{\clubsuit}$}
\newcommand{\affilumdlsc}{${}^{\diamondsuit}$}
\newcommand{\affilumdumiacs}{${}^{\heartsuit}$}
\newcommand{\affiluw}{${}^{\spadesuit}$}
\newcommand{\affilmsrny}{${}^{\nabla}$}
\author{Allyson Ettinger\affilumdling\affilumdumiacs\quad Sudha Rao\affilumdcs\affilumdumiacs \quad Hal Daum\'e III\affilumdling\affilumdcs\affilumdlsc\affilumdumiacs\affilmsrny \quad Emily M.\ Bender\affiluw \\
  University of Maryland: Linguistics\affilumdling, Computer Science\affilumdcs, Language Science\affilumdlsc and UMIACS\affilumdumiacs \\
  University of Washington Department of Linguistics\affiluw \\
  Microsoft Research New York\affilmsrny\\
  {\tt \{aetting@, raosudha@cs., hal@umiacs.\}umd.edu, ebender@uw.edu}}
\date{}

\begin{document}

\maketitle
%
%
\begin{abstract}
This paper presents a summary of the first Workshop on Building Linguistically Generalizable Natural Language Processing Systems, and the associated \textit{Build It Break It, The Language Edition} shared task. The goal of this workshop was to bring together researchers in NLP and linguistics with a shared task aimed at testing the generalizability of NLP systems beyond the distributions of their training data.  We describe the motivation, setup, and participation of the shared task, provide discussion of some highlighted results, and discuss lessons learned. 
\end{abstract}

\section{Introduction}
Machine learning techniques have had tremendously positive impact on the field of natural language processing, to the point that we now have systems for many NLP problems that work extremely well---at least when the NLP problem is carefully designed and these systems are tested on data that looks like their training data. Especially with the influx of deep learning approaches to NLP, we find ourselves more and more in the situation that we have systems that work well under some conditions, but we (and the models!)\ may have little idea what those conditions are.

We believe that linguistic knowledge is critical in many phases of the NLP pipeline, including:

\begin{enumerate}
\itemsep0em
\item\label{task} Task design and choice of language(s)
\item\label{annot} Annotation schema design
\item\label{sys} System architecture design and/or feature design
\item\label{eval} Evaluation design and error analysis
\item\label{gen} Generalization beyond training data
\end{enumerate}

Our goal in this workshop was to bring together researchers from NLP and linguistics through a carefully designed shared task. This shared task was designed to test the true generalization ability of NLP systems beyond the distribution of data on which they may have been trained. In addition to the shared task, the workshop also welcomed research contribution papers.

In this paper, we describe the shared task, laying out our motivations for pursuing this twist on the traditional set up (\S\ref{sec:brittle}) and the various design decisions we made as we took the initial idea and worked to shape it into something that would be feasible for participants and informative for our field (\S\ref{sec:task}). We then go on to describe our data (\S\ref{sec:data}), the participating systems and breaker approaches (\S\ref{sec:teams}), and our approach to scoring (\S\ref{sec:scoring}).  Finally, we give an overview of the shared task results in \S\ref{sec:results}, and discuss lessons learned in \S\ref{sec:lessons}.  

Our hope is that in laying out the successes and challenges of the first iteration of this shared task, we can help future shared tasks of this type to build on our experience. To this end, we also make available the datasets collected for and created during the shared task (\S\ref{sec:data}).


\section{Motivation: Robust NLP Systems}
\label{sec:brittle}


Natural language processing has largely embraced the ``independently and identically distributed''  (iid) probably-approximately-correct (PAC) model of learning from the machine learning community~\cite[c.f.][]{valiant1984theory}, typically under a uniform cost function. 
This model has been so successful that it often simply goes unquestioned as the ``right way'' to do NLP. Under this model, any phenomenon that is sufficiently rare in a given corpus (seen as a ``distribution of data'') is not worth addressing. Systems are not typically built to handle tail phenomena, and iid-based learning similarly trains systems to ignore such phenomena. This problem is exacerbated by frequent use of overly simplistic loss functions, which further encourage systems to ignore phenomena that they do not capture adequately.

The result is that NLP systems are quite brittle in the face of infrequent linguistic phenomena,\footnote{During a panel at the 1st Workshop on Representation Learning for NLP (ACL 2016; \url{https://sites.google.com/site/repl4nlp2016/}) some panelists acknowledged the fact that they could probably break any NLP system with very little effort---meaning it shouldn't be hard to invent reasonable examples that would confuse the systems.} a characteristic which stands in stark contrast to human language users, who at a very young age can make subtle distinctions that have little support in the distribution of data they've been exposed to~\cite[c.f.,][]{legate2002empirical,crain1987structure}. 
This ability also allows humans to avoid making certain errors due to over- or under-exposure. A computational counter-example is ignoring negations because they are relatively infrequent and typically only have a small effect on the loss function used in training. 

The brittleness of current NLP systems, and the substantial discrepancy between their capacities and that of humans, suggests that there is much left to be desired in the traditional ``iid'' model. This applies not only to training and testing, but also to error analysis: iid development data is unlikely to exhibit all the linguistic phenomena that we might be interested in testing.
Even if one is uninterested in the scientific questions addressed by testing a model's ability to handle less frequent phenomena, it should be noted that any NLP system that is released is likely to be adversarially tested by users who want to break it for fun.

This state of affairs has not gone unnoticed. On the one hand, there is work on creating targeted evaluation datasets that exhibit and are annotated for particular linguistic phenomena, in order to facilitate fine-grained analysis of the linguistic capacities of systems for tasks such as parsing, entailment, and semantic relatedness~\cite{Rim:Cla:Ste:09,Ben:Fli:Oep:11,marelli2014sick}. Additionally, there is an increasing amount of work on developing methods of exposing exactly what linguistic knowledge NLP models develop~\cite{kadar2016representation,li2015visualizing} and what linguistic information is encoded in models' produced representations~\cite{adi2016fine,ettinger2016probing}. Our aim in organizing this workshop was to build on this foundation, designing the shared task to generate data specifically created to identify the boundaries of systems' linguistic capacities, and welcoming further related research contributions to stimulate additional discussion.


\section{Shared Task: \textit{Build It Break It, The Language Edition}}

\label{sec:task}

To address the issues identified above, we developed a shared task inspired by the Build It Break It Fix It Contest\footnote{\url{https://builditbreakit.org}} and adapted for application to NLP.
The shared task proceeded in three phases: a building phase, a breaking phase, and a scoring phase:

\begin{enumerate}
\item In the first phase, ``builders'' take a designated NLP task and develop techniques to solve it.
\item In the second phase, ``breakers'', having seen the output of the builders' systems on some development data, are tasked with constructing minimal-pair test cases intended to identify the boundaries of the systems' capabilities. 
\item In the third phase, builders run their systems on the newly created minimal pair test set and provide their predictions for scoring.
\end{enumerate}

Builders are scored based how well their systems can withstand the attacks of breakers, and breakers are scored based on how well they can identify system boundaries.

The goals of this type of shared task are multi-fold: we want to build more reliable NLP technology, by stress-testing against an adversary; we want to learn more about what linguistic phenomena our systems are capable of handling so that we can guide research in interesting directions; we want to encourage researchers to think about what assumptions their models are implicitly making by asking them to break them; we want to engage linguists in the process of testing NLP systems; we want to build a test collection of examples that are not necessarily high probability under the distribution of the training data, but are nonetheless representative of language phenomena that we expect a reasonable NLP system to handle; and we want to increase cross-talk between linguistics and natural language processing researchers.

\subsection{Task Selection}
\label{sec:tasksel} 


In selecting the NLP task to be solved by the builders, we had a number of considerations. The task should be one that requires strong linguistic capabilities, so that in identifying the boundaries of the systems, breakers are encouraged to target linguistic phenomena key to increasing the robustness of language understanding. Additionally, we want the task to be without significant barrier to entry, to encourage builder participation.

In the interest of balancing these considerations and testing the effectiveness of different tasks, we ran two tasks in parallel: sentiment analysis and question-answer driven semantic role labeling~\cite[QA-SRL;][]{he2015question}.
The sentiment task consists of standard sentiment analysis performed on movie reviews. In the QA-SRL task, the input is a sentence and a question related to one of the predicates in the sentence, and the output is a span of the sentence that answers the question. See Figure~\ref{fig:qasrl} for an example item. The task allows for testing semantic role labeling without the need for a pre-defined set of roles, or for annotators with significant training or linguistic expertise.
%
\begin{figure}
\centering
\begin{tabularx}{.48\textwidth}{|l X|}\hline
\textbf{Sentence} &  UCD finished the 2006 championship as Dublin champions, by beating St Vincents in the final. \\
\textbf{Predicate} & beating \\
\textbf{Question} & Who beat someone? \\
\textbf{Answer}  &  UCD \\\hline
\end{tabularx} 
\caption{Example QA-SRL item}\label{fig:qasrl}
\end{figure}




\subsection{Building}

From the builders' point of view, the shared task is similar to other typical shared tasks in our field.  Task organizers provide training and development data, and the builder teams create systems on the basis of that data. We do not distinguish open versus closed tracks (use of provided training data is optional). Our goal was to attract a variety of approaches, both knowledge engineering-based and machine learning-based. 

We considered requiring builders to submit system code as an alternative to running their systems on two different datasets (see Section~\ref{sec:data}). However, ultimately we decided in favor of builder teams running their own systems and submitting predictions in both phases.



\subsection{Breaking}
The task of breaker teams was to construct minimal pairs to be used as test input to the builder systems, with the goal of identifying the boundaries of system capacities. In order for a test pair to be effective in identifying a system's boundaries, it needs to satisfy two requirements: 
\begin{enumerate}
\item\label{break} The system succeeds on one item of the pair but fails on the other.
\item The difference between the items in the pair is specific enough that the ability of the system to handle one but not the other can be attributed to an identifiable cause.
\end{enumerate}
Satisfaction of requirement~\ref{break} is what we will refer to as ``breaking'' a system (note that this also applies if the system fails on the original example but succeeds on the hand-constructed variant). 

Breakers were thus instructed to create minimal pairs on which they expected systems to make a correct prediction on one but not the other of the items. Breakers were additionally asked, while constructing minimal pairs, to keep in mind what exactly they would be able to conclude about a system's linguistic capacity if it proved able to handle one item of a given pair but not the other. Along this line, breakers were encouraged to provide a rationale with each minimal pair, to explain their reasoning in making a given change.\footnote{Breaker instructions can be found here: \url{https://bibinlp.umiacs.umd.edu/sharedtask.html}}

In order to exert a certain amount of control over the domain and style of breakers' items,
we required breakers to work from data provided for each task. Specifically, we asked them to select sentences from the provided dataset and make targeted changes in order to create their minimal pairs. This means that each minimal pair consisted of one unaltered sentence from the original dataset and one sentence reflecting the breakers' change to that sentence.  This was done to ensure that systems had at least a reasonable chance at success, by scoping down the range of possible variants that breakers could provide.


      
      
As an example, let us say that the provided sentiment analysis dataset includes the sentence \emph{I love this movie}, which has positive sentiment (+1). A breaker team could then construct a pair such as the following:
\ex.\label{sentex}
\begin{itemize}
\item[+1] \ \ I love this movie!
\item[+1] \ \ I'm mad for this movie!
\end{itemize}


While the first item is likely straightforward to classify, we might anticipate a simple sentiment system to fail on the second, since it may flag the word \emph{mad} as indicating negative sentiment. Breakers could choose to change the sentiment with their modification, or let it remain the same. 

For the QA-SRL task, breakers were only to change the original sentence (and, if appropriate, the answer), leaving the question unaltered. For instance, breakers could generate the following item to be paired with the example in Figure~\ref{fig:qasrl}:

\ex.
\begin{itemize}
\item[{\bf Sent$'$}] \  UCD finished the 2006 championship as Dublin champions, when they beat St Vincents in the final.
\item[{\bf Ans$'$}] \ \ UCD (unchanged)
\end{itemize}

We might anticipate that the system would now predict the pronoun \emph{they} as the answer to the question, without resolving to UCD.\footnote{Breakers were not allowed to change the sentence such that the accompanying question was no longer answerable with a substring from the original sentence. For instance, breakers could not make a change such as \emph{Terry fed Parker} $\rightarrow$ \emph{Parker was fed} with an accompanying test question of \emph{Who fed Parker?}, since the answer to that question would no longer be contained in the sentence.}

The sets of minimal pairs created by the breakers then constituted the test set of the shared task, which was sent to builders to generate predictions on for scoring. 

\section{Shared Task Data}
\label{sec:data}

\subsection{Training Data}



For the sentiment training data, we used the Sentiment Treebank dataset from Socher et al.~\shortcite{socher2013recursive}, developed from the Rotten Tomatoes review dataset of Pang and Lee~\shortcite{pang2005seeing}.\footnote{Sentiment training data available here: \url{https://nlp.stanford.edu/sentiment/}} Each sentence in the dataset has a  sentiment value between 0 and 1, as well as sentiment values for the phrases in its syntactic parse. In order to establish a binary labeling scheme at the sentence level, we mapped sentences in range (0, 0.4) to ``negative'' and sentences in range (0.6, 1.0) to ``positive''. Neutral sentences---those with a sentiment value between 0.4 and 0.6---were removed. The sentiment training data had a total of 6921 sentences and 166738 phrases. Phrase-level sentiment labels were made available to participants as an optional resource.

For QA-SRL training data, we used the data created by He et al.~\shortcite{he2015question}.\footnote{QA-SRL training data available here: \url{https://dada.cs.washington.edu/qasrl/}.} These items were drawn from Wikipedia, and each item of the training data includes the sentence (with the relevant predicate identified), the question, and the answer. The training data had a total of 5149 sentences.
\subsection{Blind Development Data}
Blind dev data was provided for builders to submit initial predictions on, as produced by their systems. These predictions were made available for breakers, to be used as a reference when creating test minimal pairs.
For sentiment, we collected an additional 500 sentences from a pool of Rotten Tomatoes reviews for movies in the years 2003-2005. For annotations, we used the same method of annotation via crowd-sourcing that was used by Socher et al.~\shortcite{socher2013recursive}. 
For QA-SRL, we extracted a set of 814 sentences from Wikipedia and annotated these by crowd-sourcing, following the method of He et al.~\shortcite{he2015question}.


\subsection{Starter Data for Breakers}
As described above, breakers were given data from which to draw items that could then be altered to create minimal pairs.
Sentiment breakers were provided an additional set of 500 sentiment sentences, collected and annotated by the same method as that used for the 500 blind dev sentences for sentiment.
QA-SRL breakers were provided an additional set of 814 items, collected and annotated by the same method as the blind dev items for QA-SRL.


\subsection{Test Data}
The test data for evaluating builder systems consisted of the minimal pairs constructed by the breaker teams. The labels for the pairs were provided by the breakers themselves, though additional crowd-sourced labels were made available for teams to check for any substantial deviations.\\

We release the minimal pair test sets, as well as annotated blind dev and starter data for sentiment and QA-SRL: \url{https://bibinlp.umiacs.umd.edu/data}.

\section{Task Participants}
\label{sec:teams}

\subsection{Builder Teams: Sentiment}
\paragraph{Strawman}
Kyunghyun Cho contributed a sentiment analysis system intended to serve as a na{\"\i}ve baseline for the shared task. This model, called Strawman, consisted of an ensemble of five deep bag-of-ngrams multilayer perceptron classifiers. The model's vocabulary was composed of the most frequent 100k n-grams from the provided training data, with \emph{n} up to 2~\cite{cho2017recursive}. 
\paragraph{University of Melbourne, CNNs}
The builder team from University of Melbourne (which also participated as a breaker team), contributed two sentiment analysis systems consisting of convolutional neural networks. One CNN was trained on data labeled at the phrase level (PCNN), and the other was trained on data labeled at the sentence level (SCNN)~\cite{li2017bibi}.  

\paragraph{Recursive Neural Tensor Network }

To supplement our submitted builder systems, we tested several additional sentiment analysis systems on the breaker test set. The first of these was the Stanford Recursive Neural Tensor Network (RNTN)~\cite{socher2013recursive}. This model is a recursive neural network-based sentiment classifier,
composing words and phrases of input sentences based on binary branching syntactic structure, and using the composed representations as input features to softmax classifiers at every syntactic node. This model, rather than parameterizing the composition function by the words being composed~\cite{socher2012semantic}, uses a single more powerful tensor-based composition function for composing each node of the syntactic tree. 

\paragraph{DCNN}
The second supplementary sentiment system was the Dynamic Convolutional Neural Network from University of Oxford~\cite{kalchbrenner2014convolutional}. This is a convolutional neural network sentiment classifier that uses interleaved one-dimensional convolutional layers and dynamic k-max pooling layers, and handles input sequences of varying length.

\paragraph{Bag-of-ngram features}
Finally, we tested an additional bag-of-ngrams sentiment system with \emph{n} up to 3, consisting of a linear classifier, implemented by one of the organizers in vowpal wabbit~\cite{langford2007vowpal}. 

\subsection{Breaker Teams: Sentiment}

\paragraph{Utrecht} The breaker team from Utrecht University used a variety of strategies, including insertion of modals and opinion adverbs that convey speaker stance, changes based in world knowledge, and pragmatic and syntactic changes~\cite{staliunaite2017breaking}. 
\paragraph{Ohio State University} The breaker team from OSU also used a variety of strategies, classified as morphosyntactic, semantic, pragmatic, and world knowledge-based changes, to target hypothesized weaknesses in the sentiment analysis systems~\cite{mahler2017breaking}. 
\paragraph{University of Melbourne} The breaker team from University of Melbourne opted to generate test minimal pairs automatically,  borrowing from methods for generating adversarial examples in computer vision. They used reinforcement learning, optimizing on reversed labels, to identify tokens or phrases to be changed, and then applied a substitution method~\cite{li2017bibi}. Some human supervision was used to ensure grammaticality and correct labeling of the sentences.
\paragraph{VTeX company, Vilnius} The fourth sentiment breaker team did not submit a paper, but provided us with the following system description: ``The basic approach of breaking strategy was to use sequence2sequence deep learning approach. We used whole Movie review data\footnote{http://www.cs.cornell.edu/people/pabo/movie-review-data/scale\_whole\_review.tar.gz} for the first step of the training and BIBI training data for the second step of the training. We used OpenNMT toolkit for implementing seq2seq strategy. The training of deep neural network was done in two steps. First, the neural network was trained to replicate the given sentence from Movie review data corpus with 4 layers and 1000 nodes. On GPU it took 5 hours. Second, we used BIBI training data to adjust the trained model. The idea was to use three BIBI datasets: whole, positive only, negative only. So, the trained model was retrained to replicate sentences of these three subsets separately. This gave three different trained models. The hypothesis was that, for the given test sentence, neural network model will be able to generate more likely positive sentence with the positive model, more likely negative sentence with the negative model, and some variation with the general model. We gave test sentences to these three model to generate new sentences. Half of the generated sentences had UNK words. So, these sentences were discarded. Many generated sentences had no changes. There was about 150 sentences that had changes and we manually selected 100 sentences for submission that had meaningful changes.''

\subsection{Builder Team: QA-SRL}
The organizers provided a QA-SRL system, as there were no external builder submissions for this task. The provided system was a logistic regression classifier, trained with 1-through-5 skip-grams with a maximum skip of 4. Potential answers were neighbors and neighbors-of-neighbors in a dependency parse of the sentence~\cite[Stanford dependency parser;][]{de2006generating}, and input to the classifier was the predicate,  question verb, question string, and dependency relation between the predicate and the potential answer. An answer was marked as correct at training time if it overlapped at least 75\% in characters with the true answer.

\subsection{Breaker Team: QA-SRL}
There was one breaker submission for QA-SRL from UMass Amherst consisting of Carolyn Anderson, Su Lin Blodgett, Abram Handler, Katie Keith, Brendan O'Connor. This team did not submit a paper but provided us with the following system description:
``Our team used a wide variety of linguistically motivated strategies to confuse natural language processing models, including, for example, adding neither/nor constructions, adding coordination, changing prepositions, shifting word senses, increasing token distances, changing word order, adding additional dates and times, pronoun resolution, etc. Individual annotators labeled examples with brief explanations, which are available in our data spreadsheet\footnote{https://go.umd.edu/bibinlp-qasrl-breaker-umass}''

\section{Shared Task Scoring}
\label{sec:scoring}

For the purpose of scoring, a test minimal pair is considered to have ``broken'' a system if one item of the pair gets a correct prediction and the other item gets an incorrect prediction. As outlined above, this is to reward breakers for zeroing in on system boundaries. 

For scoring the breakers, we decided to use the average across systems of the product of the system dev set accuracy and system breaking percentage. 
Specifically, if a breaker $j$ provides a set of examples $D_j$ to break systems $i=1 \dots N$, then the breaker score is: 
\begin{align}
\text{score}(j) &=\frac 1 N \sum_{i=1}^N \text{acc}_i(\text{dev}) \frac {\text{break}(i,j)} {|D_j|} \\
\text{acc}_i(\text{dev}) &= \text{accuracy of system $i$ on dev} \\
\text{break}(i,j) &= \# x \in D_j \text{ that break system } i
\end{align}
%
The motivation here is to weight breaker successes against a given system by the general strength of that system.

For scoring the builders, we used two metrics: 
\begin{enumerate}
\itemsep0em
\item Average F score across all sentences (originals and modified) for all breaker teams
\item Percentage of sentence pairs that break system.
\end{enumerate}

\begin{figure*}[ht]
\begin{center}
\frame{\includegraphics[width =.8\textwidth]{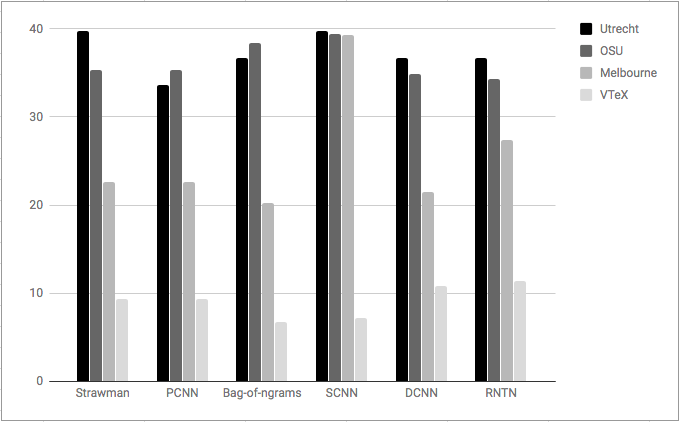}}
\end{center}
\caption{Detailed breaking percentages}\label{fig:detailed}
\end{figure*}

\section{Results and Discussion} 
\label{sec:results}




Since our participation in the QA-SRL task was minimal, we focus in this section on the results for the sentiment analysis task.

\subsection{Aggregate Results}

\begin{table}[ht]
\centering
\begin{tabular}{lcc}
\toprule
                & \textbf{average} & \textbf{\% broken} \\
\textbf{System} & \textbf{F1}      & \textbf{test cases} \\
\midrule
Strawman & \textbf{0.528} & 25.43 \\
Phrase-based CNN & 0.518 & \textbf{24.39}  \\ 
Bag-of-ngrams & 0.510 & 24.74 \\
Sentence-based CNN & 0.490 & 28.57  \\ 
DCNN & 0.483 & 25.09  \\ 
RNTN & 0.457 & 25.96  \\ 
\bottomrule
\end{tabular}
\caption{Builder team scores: Average F1 across all breaker test cases, and percent of breaker test cases that broke the system}\label{tab:builder}
\end{table}

\begin{table}[ht]
\centering
\begin{tabular}{lc}
\toprule
\textbf{Breaker} & \textbf{score}  \\
\midrule
Utrecht & \textbf{31.17}  \\
OSU & 28.66  \\
Melbourne & 19.28  \\ 
VTeX & 7.48  \\ 
\bottomrule
\end{tabular}
\caption{Breaker team scores}\label{tab:breaker}
\end{table}

Aggregate results for builders are shown in Table~\ref{tab:builder}. Computing by F1 score, Strawman comes out on top among builder systems with an average F1 of 0.528, followed by the phrase-based CNN and bag-of-ngrams. When scored by percentage of pairs that break the system, the phrase-based CNN comes out on top, broken by 24.39\% of test pairs. The bag-of-ngrams model and DCNN follow closely behind, while the sentence-based CNN falls last by a fair margin. 



Aggregate results for breakers are shown in Table~\ref{tab:breaker}. By our chosen scoring metric, the team from Utrecht falls in first place among breaker teams, followed closely by the breaker team from OSU.


\subsection{Detailed Results}
Aggregate scores obscure the important details that we aim to probe for with this shared task, namely the particular weaknesses of a given system targeted by a given minimal pair or set of minimal pairs. Figure~\ref{fig:detailed} brings us closer to the desired granularity with individual breaking percentages, allowing us a clearer sense of the interaction between breaker team and builder system. 

Some patterns emerge. The Utrecht and OSU breaker team are roughly on par across systems, with Utrecht pulling ahead by the largest margin on Strawman. These teams seem to have used a comparable variety of linguistically diverse and targeted attacks, which may explain the fact that they perform similarly.

The Melbourne test set stood out from the others in that it was automatically generated. As might be expected, this test set lags behind in breaking percentage against most systems---however, against the sentence-based CNN it performs on par with the other two teams. 

The VTeX test set has the lowest overall breaking percentages by a substantial margin. One interesting note is that this team's test set receives one of its lowest breaking percentages against the sentence-based CNN, which was the source of some of the highest breaking percentages for the other breaker teams. 

\subsection{Item-based Results}

It is of course at the level of individual minimal pairs that our analysis of this shared task can have the most power. Tables~\ref{tab:minipairs} and~\ref{tab:mppreds} show a sample of breaker minimal pairs and builder system predictions on those pairs, allowing us to observe system performance at the item level for this sample. These examples were chosen with the goal of finding interesting strategies that break some systems but not others, in order to explore differences. However, we found that for a majority of successful test pairs, systems tended to break together.

On Utrecht pair 1a/b, we see that the addition of the word \emph{pain} breaks Strawman and bag-of-ngrams, as we might expect from ngram-based systems. Apart from RNTN, which makes incorrect predictions on both items, the remaining systems are able to handle this change.

On Utrecht pair 2a/b, we see that bag-of-ngrams, DCNN, SCNN and RNTN all break, though in different directions, with SCNN and RNTN getting the altered sentence wrong, and bag-of-ngrams and DCNN getting the original sentence wrong. This suggests a lack of sensitivity to the subtly different sentiments conveyed in context by the substituted words \emph{unnerving} and \emph{hilarious}. Strawman and PCNN, however, predict both items correctly.

The substitution of the comparative phrase in OSU pair 1a/b impressively breaks every system, suggesting that the sentiment conveyed by the phrase \emph{just willing enough} in context is beyond the capacity of any of the systems. The sarcasm addition in OSU 2a/b breaks Strawman, bag-of-ngrams and DCNN, but not SCNN or RNTN (while PCNN breaks in the opposite direction).

Strawman breaks on Melbourne 1a/b, which is interesting as we might expect the substituted item \emph{thrill} to be flagged as carrying positive sentiment. Bag-of-ngrams fails on both items of the pair, and RNTN gives a neutral label for the second item.

Melbourne 2a/b employs a word re-ordering technique and breaks every system in various directions---except for bag-of-ngrams and RNTN, which fail on both items---suggesting that both the original and altered sentences of this pair give systems trouble.  

VTeX 1a/b fools bag-of-ngrams with the altered sentence, while DCNN and RNTN make incorrect predictions on the original.

As we can see in these examples, by testing systems on minimal pair test items such as these we have the potential to zero in on the linguistic phenomena that any given system can and cannot handle. It is also clear that it is specifically when a system ``breaks'' (makes a correct prediction on one but not the other item), and when the change in the pair is targeted enough, that we are able to draw straightforward conclusions. For instance, OSU pair 1a/b allows us to conclude that inferring the positive effect of the phrase \emph{just [...] enough} on a previously negative context is beyond the systems' capacities. On the other hand, the more diffuse changes in Melbourne pair 2a/b make it more difficult to determine the precise cause of a system breaking in one direction or the other. 

Of course, to be more confident about our conclusions, we would want to analyze system predictions on multiple different pairs that target the same linguistic phenomenon. This can be a goal for future iterations and analyses.



\begin{table*}
\begin{tabularx}{\textwidth}{ c | p{9cm} | x{1cm} | p{2.3cm} }
ID & Minimal Pairs & Label & Rationale  \\ \hline 

\multirow{2}{*}{Utrecht 1a} & Through elliptical and seemingly oblique methods, he forges moments of staggering \textbf{emotional power} & \multirow{2}{*}{+1} & \multirow{4}{2.3cm}{\textit{Emotional pain can be positive}}\\ 
\multirow{2}{*}{Utrecht 1b} & Through elliptical and seemingly oblique methods, he forges moments of staggering \textbf{emotional pain} & \multirow{2}{*}{+1} & \\ \hline

\multirow{2}{*}{Utrecht 2a} & [Bettis] has a smoldering, humorless intensity that's \textbf{unnerving}.
& \multirow{2}{*}{-1} & \multirow{4}{2.3cm}{\textit{Funny can be positive \& negative}}  \\ 
\multirow{2}{*}{Utrecht 2b} & [Bettis] has a smoldering, humorless intensity that's \textbf{hilarious}.
& \multirow{2}{*}{+1} &  \\ \hline
\hline

\multirow{2}{*}{OSU 1a} & A bizarre (and sometimes repulsive) exercise that's \textbf{a little too willing} to swoon in its own weird embrace. & \multirow{2}{*}{-1} & \multirow{4}{2.3cm}{\textit{Comparative}}\\ 
\multirow{2}{*}{OSU 1b} & A bizarre (and sometimes repulsive) exercise that's \textbf{just willing enough} to swoon in its own weird embrace.  & \multirow{2}{*}{+1} & \\ \hline
 
 \multirow{2}{*}{OSU 2a} & Proves that \textbf{fresh new work} can be done in the horror genre if the director follows his or her own shadowy muse. & \multirow{2}{*}{+1} & \multirow{4}{2.3cm}{\textit{Sarcasm} \textit{(single cue)}} \\ 
\multirow{2}{*}{OSU 2b} & Proves that \textbf{dull new work} can be done in the horror genre if the director follows his or her own shadowy muse. & \multirow{2}{*}{-1} & \\ \hline
 \hline
 
 \multirow{2}{*}{Melbourne 1a} & Exactly the kind of \textbf{unexpected delight} one hopes for every time the lights go down. & \multirow{2}{*}{+1} & \multirow{4}{2.3cm}{\textit{(Not provided)}}  \\ 
\multirow{2}{*}{Melbourne 1b} & Exactly the kind of \textbf{thrill} one hopes for every time the lights go down. & \multirow{2}{*}{+1} & \\ \hline
 
 \multirow{2}{*}{Melbourne 2a} & \textbf{American drama} doesn't get any more meaty and muscular \textbf{than this}.  & \multirow{2}{*}{+1} & \multirow{4}{2.3cm}{\textit{(Not provided)}}  \\ 
\multirow{2}{*}{Melbourne 2b} & \textbf{This} doesn't get any more meaty and muscular \textbf{than American drama}. & \multirow{2}{*}{-1} \\ \hline
 \hline
 
  \multirow{2}{*}{VTeX 1a} & Rarely have good intentions been wrapped in such a \textbf{sticky} package. & \multirow{2}{*}{-1} & \multirow{4}{2.3cm}{\textit{(Not provided)}} \\ 
\multirow{2}{*}{VTeX 1b} & Rarely have good intentions been wrapped in such a \textbf{adventurous} package.  & \multirow{2}{*}{+1}  \\ \hline
 
\end{tabularx}
\caption{\textbf{Sample minimal pairs}: Examples of minimal pairs created by different breaker teams with the minimal changes highlighted. `Label' is the label provided to the pairs by the breaker teams. }\label{tab:minipairs}
\end{table*} 

\begin{table*}
\begin{tabular}{c | c | c | c | c | c | c | c}
ID & True Label & Strawman &
PCNN & Bag-of-ngrams & SCNN & DCNN & RNTN  \\ \hline 
Utrecht 1a  & +1 & +1 & +1 & +1 & +1 & +1 & -1 \\ 
Utrecht 1b  & +1 & -1 & +1 & -1 & +1 & +1 & -1 \\ \hline
\hline

Utrecht 2a  & -1 & -1 & -1 & +1 & -1 & +1 & -1 \\ 
Utrecht 2b  & +1 & +1 & +1 & +1 & -1 & +1 & -1 \\ \hline

OSU 1a  & -1 & -1 & -1 & -1 & -1 & -1 & -1 \\ 
OSU 1b  & +1 & -1 & -1 & -1 & -1 & -1 & -1 \\ \hline

OSU 2a  & +1 & +1 & -1 & +1 & +1 & +1 & +1 \\ 
OSU 2b  & -1 & +1 & -1 & +1 & -1 & +1 & -1 \\ \hline
\hline

Melbourne 1a  & +1 & +1 & +1 & -1 & +1 & +1 & +1 \\ 
Melbourne 1b  & +1 & -1 & +1 & -1 & +1 & +1 & 0 \\ \hline

Melbourne 2a  & +1 & -1 & +1 & -1 & -1 & -1 & -1 \\ 
Melbourne 2b  & -1 & -1 & +1 & +1 & -1 & -1 & 0 \\ \hline
\hline

VTeX 1a  & -1 & -1 & -1 & -1 & -1 & +1 & +1 \\ 
VTeX 1b  & +1 & +1 & +1 & -1 & +1 & +1 & +1 \\ \hline
\hline

\end{tabular}
\caption{\textbf{Sample minimal pair predictions}: Builder system predictions on the example minimal pairs from Table~\ref{tab:minipairs}. `True Label' is the label provided to the pairs by the breaker teams.}\label{tab:mppreds}
\end{table*}

\section{Lessons for the Future}\label{sec:lessons}
A variety of lessons came out of the shared task, which 
can be helpful for future iterations or future shared tasks of this type. We describe some of these lessons here.

The choice of NLP task is an important one. While QA-SRL is a promising task in terms of requiring linguistic robustness, it yielded lower participation than sentiment analysis. Strategies for encouraging buy-in from both builders and breakers will be important. One strategy would be to team up with existing shared tasks, to which we could add a breaking phase.

While going through the labels assigned to the minimal pairs by breaker teams, we find some label choices to be questionable. Since unreliable labels will skew the assessment of builder performance, in future iterations there should be an additional phase in which we validate breaker labels with an external source (e.g., crowd-sourcing). To minimize cost and time, this could be done only for examples that are ``contested'' by either builders or other breakers.

The notion of a ``minimal pair'' is critical to this task, so it is important that we define the notion clearly, and that we ensure that submitted pairs conform to this definition. Reviewing breaker submissions, we find that in some cases breakers have significantly changed the sentence, in ways that may not conform to our original expectations. In future iterations, it will be important to have clear and concrete definitions of minimal pair, and it would also be useful to have some external review of the pairs to confirm that they are permissible. 

For this year's shared task we chose to limit breakers by requiring them to draw from existing data for creating their pairs. A potential variation to consider would be allowing breaker teams to create their own sentence pairs from scratch, in addition to drawing from existing sentences (with the restriction that sentences should fall in the specified domain). This greater freedom for breakers may increase the range of linguistic phenomena able to be targeted, and the precision with which breakers can target them.


Finally, it is important to consider general strategies for encouraging participation. We identify two potential areas for improvement. First, the timeline of this year's shared task was shorter than would be optimal, which placed an undue burden in particular on builders, who needed to run systems and submit predictions in two different phases. A longer timeline could make participation more feasible. Second, participants may be reluctant to submit work to be broken---to address this, we might consider anonymous system submissions in the future. 


\section{Conclusion}
The First Workshop on Building Linguistically Generalizable NLP systems, and the associated first iteration of the \emph{Build It Break It, The Language Edition} shared task, allowed us to begin exploring the limits of current NLP systems with respect to specific linguistic phenomena, and to extract lessons to build on in future iterations or future shared tasks of this type. We have described the details and results of the shared task, and discussed lessons to be applied in the future. We are confident that tasks such as this, that emphasize testing the effectiveness of NLP systems in handling of linguistic phenomena beyond the training data distributions, can make significant contributions to improving the robustness and quality of NLP systems as a whole.

\section*{Acknowledgments}
The authors would like to acknowledge Chris Dyer who, as a panelist at the Workshop on Representation Learning for NLP, made the original observation about the brittleness of NLP systems which led to the conception of the current workshop and shared task. We would also like to thank the UMD Computational Linguistics and Information Processing lab, and the UMD Language Science Center. This work was partially supported by NSF grants NRT-1449815 and IIS-1618193, and an NSF Graduate Research Fellowship to Allyson Ettinger under Grant No. DGE 1322106. Any opinions, findings, conclusions, or recommendations expressed here are those of the authors and do not necessarily reflect the view of the
sponsor(s).

\bibliography{blgnlp}
\bibliographystyle{emnlp_natbib}

\end{document}